\documentclass{article}

\usepackage{graphicx}
\usepackage{mathtools}
\usepackage{amssymb}
\usepackage{amsmath}
\usepackage{float}

\newcommand{\vc}[1]{\ensuremath{\mathbf{#1}}}

\newcommand{\inv}{^{-1} }
\newcommand{\sbr}[1]{\ensuremath{_{\mathrm{#1}}}}
\newcommand{\spr}[1]{\ensuremath{^{\mathrm{#1}}}}
    \PassOptionsToPackage{numbers, compress}{natbib}

\usepackage[final]{neurips_2019_ml4ps}

\usepackage[utf8]{inputenc} 
\usepackage[T1]{fontenc}    
\usepackage{booktabs}       
\usepackage{amsfonts}       
\usepackage{microtype}      

\title{Bayesian Inversion Of Generative Models For Geologic Storage Of Carbon Dioxide}

\author{%
  Gavin H. Graham, Yan Chen\\
  Total \\
  Aberdeen, Scotland, UK. \\
  \texttt{$\left\{\text{gavin.graham, yan.chen}\right\}$@total.com} \\
}

\begin{document}

\maketitle

\begin{abstract}
Carbon capture and storage (CCS) can aid decarbonization of the atmosphere to limit further global temperature increases. A framework utilizing unsupervised learning is used to generate a range of subsurface geologic volumes to investigate potential sites for long-term storage of CO\sbr{2}. Generative adversarial networks are used to create geologic volumes, with a further neural network used to sample the posterior distribution of a trained Generator conditional to sparsely sampled physical measurements. These generative models are further conditioned to  historic dynamic fluid flow data through Bayesian inversion to improve the resolution of the forecast of the storage capacity of injected CO\sbr{2}.
\end{abstract}

\section{Introduction}
Limiting global warming to well below 2$^{\circ}$C, but preferably limiting the temperature increase to 1.5$^{\circ}$C above pre-industrial levels, was the key outcome from the 2015 United Nations International Climate Change Conference \cite{cop21}. While this requires a substantive effort in reducing emissions from energy generation and other industrial activities, removing carbon dioxide (CO\sbr{2}) from the atmosphere is also necessary to ensure that limit is not exceeded \cite{schrag07, harper18}. One method to aid decarbonization of the atmosphere is through the use of carbon capture and storage (CCS) \cite{thomas05}. CO\sbr{2} storage by injection into deep geologic formations at depleted hydrocarbon fields has been demonstrated as a safe and effective method \cite{jenkins12}. To adequately assess the potential site for long-term storage of CO\sbr{2}, the subsurface geologic volume of interest must be characterized in order to provide numerical models for fluid flow simulation experiments of CO\sbr{2} injection. 
\par
\begin{figure}[!htb]
  \centering
  \includegraphics[width=\textwidth]{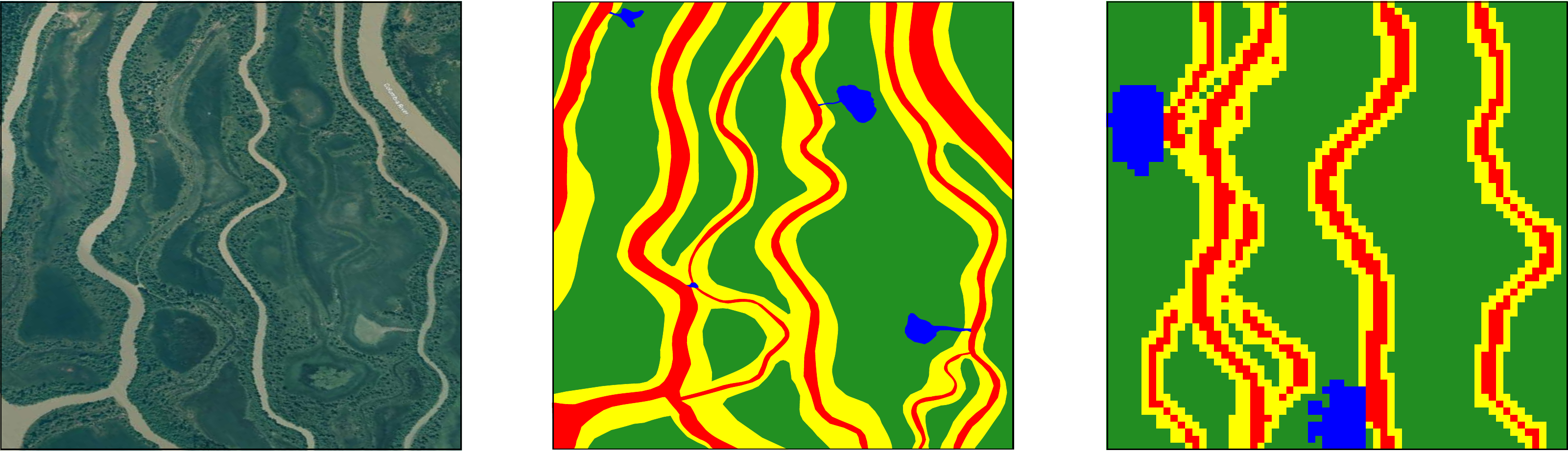}
  \caption{Left: Image of the Columbia river, British Columbia, Canada \cite{google}, which can be used as a modern analog to ancient sedimentary systems. Middle: Interpretation of sedimentary features, with river channels in red, levees in yellow and crevasse splays in blue. Right: A stochastic geostatistical simulation of the sedimentary features \cite{deutsch92}.}
  \label{fig:one}
\end{figure}

 Detailed characteristics of subsurface geologic properties are only available from sparse physical measurements, such as borehole logs which sample only a small fraction of the subsurface volume. It is therefore necessary to infer geologic properties away from these available data points. Statistical methods for creating geologic patterns and features include geostatistics \cite{deutsch92}. Such methods can be used to generate samples of the subsurface that mimic observed geologic features that are far from multivariate Gaussian random fields, but exhibit clear spatial correlations (Fig.~\ref{fig:one}). In addition to borehole data, historic fluid flow data from the depleted hydrocarbon field can be used to further calibrate the geologic models in order to reduce uncertainty in the subsurface geologic properties \cite{alfi16, mosser19}. Markov Chain Monte Carlo (MCMC) methods could be used to sample conditional realizations, but are computational costly as a significant number of fluid flow simulations are typically required \cite{laloy18}. Alternatively, ensemble-based data assimilation methods are shown to provide satisfactory approximate sampling with a limited amount of computational cost in applications ranging from weather forecasting to subsurface characterization \cite{evensen:09a}. 
\par However, geologic model parameterization is challenging for inversion techniques which rely on the Gaussian assumption \cite{canchumuni:19}. In this paper we utilize Generative Adversarial Networks (GANs) as a method for dimensionality reduction and re-parameterization of the geologic models. A pre-trained Generator is then combined with a neural network, the \textit{inference network}, that is trained to sample the posterior distribution of the latent input to the Generator, conditional to the data at known borehole locations. This preserves the parametrization of the Generator and reduces the number of computationally expensive fluid flow simulations required for uncertainty quantification and inversion problems. The Generator and Inference networks are then used to produce a range of plausible geologic models conditional to the borehore data. An ensemble-based data assimilation method is then used to obtain samples of the geologic model that also honor historic fluid flow data. Finally, the calibrated models are used to forecast the subsurface response to CO\sbr{2} injection.

\section{Methodology}
\subsection{Generative Adversarial Networks}
Generative Adversarial Networks (GANs) \cite{Goodfellow2014} consist of two competing functions (typically neural networks), a Generator $G(z)$  and a Discriminator $D(y)$, with an objective function to find the Nash equilibrium between the two networks. The Generator $G(z)$ maps an input noise vector $z$ from a simple Gaussian distribution to a synthetic sample $y$. The goal of $G(z)$ is to generate samples with a distribution $p\sbr{g}$ that is close to the data generating distribution $p_{\mathrm{data}}$. The Discriminator, $D(y)$, is a classifier that takes a sample as input and tries to determine if this sample is real or synthetic. The output of $D(y)$ is a scalar representing the probability of $y$ coming from the data. This interplay between $G$ and $D$ is formulated as a two-player minimax game:
\begin{equation}\label{eq:gan}
\min_{G}\max_{D} \mathbb{E}_{x \sim p\sbr{data}}  [\log D(x)] + \mathbb{E}_{\tilde{x}\sim{p\sbr{g}}}[\log(1-D(\tilde{x}))]
\end{equation}
where $p_{\mathrm{data}}$ is the data distribution and $p_g$ is the model distribution implicitly defined by $\tilde{x} = G(z), z \sim p(z)$. The training process consists of simultaneous application of Stochastic Gradient Descent on $D$ and $G$. Training alternates between $k$ steps of optimizing $D$ and one step of optimizing $G$. The process of training stops when $D$ is unable to distinguish $p\sbr{g}$ and $p_{\mathrm{data}}$ i.e.~$D(y) = 1/2 $ or when $p\sbr{g} = p_{\mathrm{data}}$. Due to the opposing nature of the objective function, training GANs is inherently unstable and finding stable training methods remains an open research problem. In particular, if the discriminator $D$ is optimally trained, it becomes saturated and provides no useful information for improvement of $G$. On the other hand, over-training $G$ might result in mode collapse of the generated distribution toward a single sample that $D$ always accepts. Wasserstein GAN (WGAN) \cite{arjovsky17} minimizes the Wassterstein distance rather than the Jensen-Shannon divergence in \cite{Goodfellow2014} and provides a meaningful loss metric during training that correlates with the quality of generated samples. This is important for assessing training progress and providing a criteria for convergence and was used in this study. For more detailed discussion on WGAN, the reader is referred to \cite{arjovsky17}.

\subsection{Inference network}
Generated images must be calibrated to borehole data as geologic properties are known at these locations \textit{a priori} to image generation. Given a pre-trained generator $G$, the goal is to find $z$ such that $G(z)$ honours the borehole data. An \textit{inference network}, $I$,  introduced by \cite{shing18} was used as a basis to retain the parameterization of $G$. Let $d\sbr{obs}$ denote the borehole data and $G(z)\sbr{obs}$ the generated image values at the borehole locations given $G(z)$. The loss can then be defined as:
\begin{equation}\label{eq:infloss}
\mathcal{L}(z) = \lVert {G(z)\sbr{obs}} - d\sbr{obs} \rVert^2 + \lambda  \lVert z \rVert^2
\end{equation}
where $\lambda$ = $\sigma^2$ and $\sigma$ is the measurement standard deviation. The first term in Eq.~\ref{eq:infloss} represents the error that is the difference between the observed borehole data and the generated images. The second term of Eq.~\ref{eq:infloss} ensures the distribution of $z$ remains close to the prior. For more information the reader is referred to \cite{shing18}. This method ensures that conditional samples of $G$ are generated without the need to repeatedly use a local optimizer and different initial guesses for $z$, or sampling the full posterior using Markov Chain Monte Carlo methods.

\subsection{Iterative ensemble smoother}
A particular type of the ensemble-based data assimilation methods, Ensemble Smoother with Multiple Data Assimilation (see procedure of ESMDA in Table \ref{tab:esmda}), is used to calibrate the input of the inference network to honor dynamic fluid flow data observed at borehole locations. The vector of model parameters $\vc{m}$ in Table \ref{tab:esmda} is the input of the inference network. It was shown that for linear Gaussian cases, $\alpha$ at all iterations need to satisfy the following condition, $\sum_{\ell=1}^{N\sbr{a}} 1/\alpha_{\ell}=1 $,in order to obtain the correct posterior mean and covariance at the final iteration in the limit of an infinite ensemble size \cite{emerick:13a}. An easy choice of $\alpha $ is $\alpha_{\ell}=N\sbr{a}$, often referred to as ESMDA with equal weights.
\begin{table}[!htb]
\begin{center}
\begin{tabular}{cp{12cm}}
\hline \\ [-1em]
1. & \textbf{Initialization:} generate initial ensemble, $\left\{\vc{m}_j^{0}\right\}_{j=1}^{N\sbr{e}}$, by sampling the prior distribution of model parameters. Choose number of data assimilations, $N\sbr{a}$ and coefficients $\alpha_{\ell}$ . \\ [0.2em]
2. & For $\ell =1 $ to $N\sbr{a}$:\\  [0.2em]
   & (a) \textbf{Forecast step:} run the forward model to compute the vector of simulated data 
\begin{equation}\label{eq:simd}
\vc{d}_{\text{sim},j}^{\ell} = \vc{g}(\vc{m}_j^{\ell}) \quad   \text{for} \, j=1,2,\ldots,N\sbr{e}
\end{equation} \\
& (b) Perturb the vector of historic data using 
\begin{equation}\label{eq:perturb}
\vc{d}_{\text{hist},j}^{\ell} = \vc{d}\sbr{hist}+ \sqrt{\alpha_{\ell}}\vc{C}\sbr{D}^{1/2}\vc{z}_j^{\ell} \quad  \text{for} \, j=1,2,\ldots,N\sbr{e}, \text{where} \, \vc{z}_j^{\ell} \sim \mathcal{N} (\vc{0},\vc{I}) 
\end{equation}\\  [0.2em]
& (c) \textbf{Analysis step:} update the vector of model parameters using
\begin{equation}\label{eq:esmda}
\vc{m}^{\ell+1}_j = \vc{m}^{\ell}_j + \vc{C}\sbr{md}^{\text{e}\,\ell}(\alpha_{\ell} \vc{C}\sbr{D} + \vc{C}\sbr{dd}^{\text{e}\,\ell})\inv \Bigl( \vc{d}_{\text{hist},j}^{\ell}-\vc{d}_{\text{sim},j}^{\ell} \Bigr) \quad \text{for} \, j=1,2,\ldots, N\sbr{e}
\end{equation}
\\
&  End (for) \\ \hline  
\multicolumn{2}{c}{Description of symbols} \\
\hline
Scalar & $\ell$ is the iteration index, $N\sbr{a}$ is the total number of iterations, $\alpha$ is a iteration tuning parameter (a easy choice is $\alpha_{\ell}=N\sbr{a}$ for all $\ell$s), $j$ is the index for realizations, $N\sbr{e}$ is the ensemble size, $N\sbr{d}$ is the number of data and $N\sbr{m}$ is the number of model parameters  \\[0.5em] 
Vector & $\vc{m}_j$ of length $N\sbr{m}$ is the $j$th realization of model parameters, $\vc{d}\sbr{hist}$ of length $N\sbr{d}$ is the historic data, $\vc{g}(\vc{m}_j)$ of length $N\sbr{d}$ is the simulated data from $\vc{m}_j$ ($\vc{g}(\cdot)$ is the flow equations represented by the reservoir simulator), $\vc{z}_j$ of length $N\sbr{d}$ is white noise sampled from $\mathcal{N} (\vc{0},\vc{I})$ \\[0.5em] 
Matrix & Data noise is assumed to follow multi-variate Gaussian distribution $\mathcal{N} (\vc{0},\vc{C}\sbr{D})$, where $\vc{C}\sbr{D}$ is a diagonal matrix of size $N\sbr{d} \times N\sbr{d}$ with ${\boldsymbol \sigma}\sbr{d}^2$ on its diagonal, $\vc{C}\sbr{md}\spr{e}$ is the sample covariance between realizations of $\vc{m}_j$ and $\vc{g}(\vc{m}_j)$, and $\vc{C}\sbr{dd}\spr{e}$ of size $N\sbr{d} \times N\sbr{d}$ is the sample covariance of realizations of $\vc{g}(\vc{m}_j)$. The superscript ``e'' indicates these covariances are computed from the ensemble \\ \hline
\end{tabular}
\end{center}
\caption{Algorithm of ESMDA (adapted from \cite{emerick:13a}).}
\label{tab:esmda}
\end{table}

\section{Results}
\begin{figure}[!htb]
  \centering
  \includegraphics[width=\textwidth]{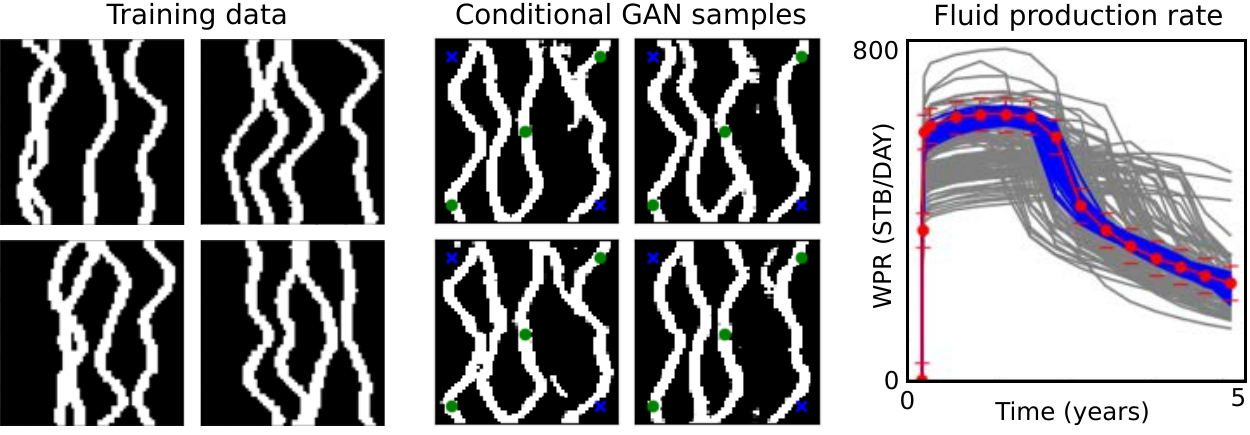}
  \caption{Left: Stochastic geostatistical simulations of simplified river channels used as training data for a GAN. Middle: Random samples from a trained Generator conditional on borehole data and historic fluid production data. Green circles represent boreholes that intersect river channels and blue crosses represent boreholes that intersect the background. Note that each conditional sample respects the known data points. Right: Fluid flow simulation data. Simulated data from initial set of generated models prior to inversion are shown in grey, and simulated data after Bayesian inversion shown in blue. Historic fluid flow data are shown as red dots with their associated uncertainty shown as error bars.}
\label{fig:two}
\end{figure}

As a first step, the methodology was tested on simplified binary representations of the geologic features in Figure \ref{fig:one}. A reference case was generated, from which borehole data were sampled and historic fluid flow data were simulated using a multi-phase fluid flow simulator. The reference geologic features were not part of the training dataset but were generated using the same parameters through a geostatistical simulation method. The images produced by the conditional Generator were geologically consistent in comparison to the training data upon visual inspection and respected the borehole data (Fig.~\ref{fig:two}). An ensemble with 100 members conditional to the borehole data was generated, followed by 20 iterations of ESMDA. In each ESMDA iteration 100 flow simulations were completed in parallel. The final calibrated models were obtained after 2000 fluid flow simulations. The match to historic fluid flow data was satisfactory (last panel of Fig.~\ref{fig:two}). 
Four realizations from the final ensemble of 100 calibrated models are shown in the middle panel of Fig.~\ref{fig:two}. Due to the constraints imposed by the borehole data and the time series of phase rates and pressure data, the final realizations share many similar features. Although there are still subtle variations in the local positioning of river channel features among the final realizations, it is very likely that the final ensemble of calibrated realizations under-estimates the true uncertainty. The input space of the inference network $I$ and the latent space of the generator $z$ also become correlated as a result of conditioning to both static and dynamic data.
%

\section{Conclusions}
Subsurface geologic volumes can be calibrated to known data in order to reduce uncertainty in numerical models for computational fluid flow experiments of CO\sbr{2} injection. Conditional sampling of a Generator network through an inference network produced geologically consistency features while honoring known static borehole data. 
The ensemble-based data assimilation method was effective in calibrating the input of the inference network to provide further conditioning to dynamic fluid flow observations. The proposed methodology can be extended to more complex and realistic geologic models and can provide a basis for assessing the capacity of CO\sbr{2} injection and storage in subsurface geologic formations.



\end{document}